\documentclass[runningheads]{llncs}

\usepackage{amsmath}

\begin{document}

\pagestyle{headings}

\mainmatter

\title{A Cross-Conformal Predictor for \\Multi-label Classification}


\author{Harris Papadopoulos}


\institute{Computer Science and Engineering Department, Frederick University,\\
7 Y. Frederickou St., Palouriotisa, Nicosia 1036, Cyprus
\email{h.papadopoulos@frederick.ac.cy}}

\maketitle

\begin{abstract}
Unlike the typical classification setting where each instance is associated with a single class, in 
multi-label learning each instance is associated with multiple classes simultaneously. Therefore the 
learning task in this setting is to predict the subset of classes to which each instance belongs. 
This work examines the application of a recently developed framework called Conformal Prediction (CP) 
to the multi-label learning setting. CP complements the predictions of machine learning algorithms 
with reliable measures of confidence. As a result the proposed approach instead of just predicting the 
most likely subset of classes for a new unseen instance, also indicates the likelihood of each predicted 
subset being correct. This additional information is especially valuable in the multi-label setting where 
the overall uncertainty is extremely high.
\end{abstract}

\section{Introduction}

Most machine learning research on classification deals with problems in which each instance is associated with 
a single class $y$ from a set of classes $\{Y_1, \dots, Y_c\}$. As opposed to this standard setting, in 
\emph{multi-label classification} each instance can belong to multiple classes, so that each instance is associated with 
a set of classes $\psi \subseteq \{Y_1, \dots, Y_c\}$, called a \emph{labelset}. 
There are many real-world problems in which 
such a setting is natural. For instance in the categorization of news articles an article discussing the positions 
of political parties on the educational system of a country can be classified as both politics and education. 
Although until recently the main multi-label classification application was the categorization of textual data, 
in the last few years an increasing number of new applications started to attract the attention of more researchers 
to this setting. Such applications include the semantic annotation of images and videos, the categorization of music 
into emotions, functional genomics, proteomics and directed marketing.

As a result of the increasing attention to the multi-label classification setting, many new machine learning techniques 
have been recently developed to deal with problems of this 
type, see e.g.~\cite{Elisseeff:akernel,zhang:mlrbf,zhang:mlnb,zhang:bpmll,zhang:mlknn}. 
However, like most machine learning methods, these 
techniques do not produce any indication about the likelihood of each of their predicted labelsets being correct. Such an 
indication though can be very helpful in deciding how much to rely on each prediction, especially since the certainty 
of predictions may vary to a big degree between instances. To address this problem this paper examines the 
application of a recently developed framework for providing reliable confidence measures to predictions, 
called \emph{Conformal Prediction} (CP) \cite{vovk:alrw}, to the multi-label classification setting. 
Specifically it follows the newly proposed \emph{Cross-Conformal Prediction} (CCP)~\cite{vovk:ccp} version of the 
framework, which allows it to overcome the 
prohibitively large computational overhead of the original CP. The proposed approach computes a p-value for each of 
the possible labelsets, which can be used either for accompanying each prediction with confidence measures that 
indicate its likelihood of being correct, or for producing sets of labelsets that are guaranteed to contain the 
true labelset with a frequency equal to or higher than a required level of confidence.

In the remaining paper, a description of the general idea behind CP and of the CCP version of the framework, is 
first provided in Section~\ref{sec:cp}. 
Section \ref{sec:nm} defines the proposed approach while Section \ref{sec:res} presents 
the experiments performed and the obtained results. Finally Section \ref{sec:conc} gives the conclusions of this 
work.

\section{Conformal and Cross-Conformal Prediction}
\label{sec:cp}

Typically in classification we are given a set of training examples $\{z_1, \dots, z_l\}$, 
where each $z_i \in \mathcal{Z}$ is a pair $(x_i,y_i)$ consisting of a vector of attributes $x_i\in \bbbr^d$ 
and the classification $y_i \in \{Y_1, \dots, Y_c\}$. We are also given a new unclassified 
example $x_{l+1}$ and our task is to state something about our confidence in each possible 
classification of $x_{l+1}$ without assuming anything more than that 
all $(x_i,y_i)$, $i=1,2, \dots,$ are independent and identically 
distributed.

The idea behind CP is to assume every possible classification $Y_j$ 
of the example $x_{l+1}$ and check how likely it is that the extended set 
of examples
\begin{equation}
\label{eq:extset}
\{(x_1, y_1), \dots, (x_l, y_l), (x_{l+1}, Y_j)\}
\end{equation}
is i.i.d. This
in effect will correspond to the likelihood of $Y_j$ being the true 
label of the example $x_{l+1}$ since this is the only unknown value 
in~(\ref{eq:extset}). 

First a function $A$ called \emph{nonconformity measure} is used to map each pair $(x_i,y_i)$ in (\ref{eq:extset}) to 
a numerical score 
\begin{subequations}\label{eq:nmdef}
\begin{align}
\alpha_i &= A(\{(x_1, y_1), \dots, (x_l, y_l), (x_{l+1}, Y_j)\}, (x_i,y_i)), \mspace{15 mu} i = 1, \dots, l, \\
\alpha^{Y_j}_{l+1} &= A(\{(x_1, y_1), \dots, (x_l, y_l), (x_{l+1}, Y_j)\}, (x_{l+1}, Y_j)),
\end{align}
\end{subequations}
called the \emph{nonconformity score} of instance $i$. This score indicates 
how nonconforming, or strange, it is for $i$ to belong in~(\ref{eq:extset}). In effect the nonconformity 
measure is based on a conventional machine learning algorithm, called the \emph{underlying algorithm} of 
the corresponding CP and measures the degree of disagreement between the actual label $y_i$ and 
the prediction $\hat y_i$ of the underlying algorithm, after being trained on~(\ref{eq:extset}). 
The nonconformity measure for multi-label learning used in this work is defined in Section \ref{sec:nm}.

The nonconformity score $\alpha^{Y_j}_{l+1}$ is then compared to the nonconformity 
scores of all other examples to find out how unusual $(x_{l+1}, Y_j)$ is according 
to the nonconformity measure used. This comparison is performed with the function
\begin{equation}
\label{eq:pvalue}
  p((x_1, y_1), \dots, (x_l, y_l), (x_{l+1}, Y_j)) = \frac{|\{i = 1, \dots, l, l+1 : \alpha^{Y_j}_i \geq \alpha^{Y_j}_{l+1}\}|}{l+1},
\end{equation} 
the output of which is called the p-value of $Y_j$, also denoted as $p(Y_j)$. An 
important property of (\ref{eq:pvalue}) 
is that $\forall \delta\in [0, 1]$ and for all probability distributions $P$ on $\mathcal{Z}$,
\begin{equation}
\label{eq:validity}
  P^{l+1}\{((x_1,y_1), \dots, (x_l, y_l), (x_{l+1},y_{l+1})):p(y_{l+1}) \leq \delta\}\leq \delta;
\end{equation}
a proof can be found in \cite{vovk:alrw}. According to this property, if $p(Y_j)$ is under some 
very low threshold, say $0.05$, this means that $Y_j$ is highly unlikely 
as the probability of such an event is at most $5\%$ if~(\ref{eq:extset})
is i.i.d.

There are two standard ways to use the p-values of all possible classifications for producing the 
output of a CP:
\begin{itemize}
\item Predict the classification with the highest p-value and output one minus the second highest p-value 
as confidence to this prediction and the p-value of the predicted classification (i.e. the highest p-value) 
as credibility.
\item Given a confidence level $1 - \delta$, output the prediction set $\{ Y_j : p(Y_j) > \delta \}$.
\end{itemize}
In the first case, confidence is an indication of 
how likely the prediction is of being correct compared to all other possible classifications, 
whereas credibility indicates how suitable the training set is for the particular instance; 
specifically a very low credibility value indicates that the particular instance does not 
seem to belong to any of the possible classifications. In the second case, the prediction set 
will not contain the true label of the instance with at most $\delta$ probability.

The important drawback of the above process is that since the last example in~(\ref{eq:extset}) 
changes for every possible classification, the underlying algorithm needs to be trained $c$ times. 
Moreover the whole process needs to be repeated for every test example. This makes it extremely 
computationally inefficient in many cases and especially in a multi-label setting where there are 
$2^n$ possible labelsets for $n$ classes, or $2^n - 1$ if we exclude the empty labelset.

To overcome this computational inefficiency problem an inductive version of the framework was 
proposed in~\cite{papa:icm-rr} and~\cite{papa:icm-pr} called \emph{Inductive Conformal Prediction} (ICP).
ICP is based on the same theoretical foundations described above, but follows a modified version of the approach, 
which allows it to train the underlying algorithm only once.
This is achieved by dividing the training set into two smaller sets, the \emph{proper training set} 
and the \emph{calibration set}. The proper training set is then used for training the 
underlying algorithm of the ICP and only the examples in the calibration set are used for calculating
the p-value of each possible classification for every test example. 

Although ICP is much more computationally efficient than the original CP approach, the fact that it 
does not use the whole training set for training the underlying algorithm and for calculating its 
p-values results in lower informational efficiency. That is the resulting prediction sets might be 
larger than the ones produced by the original CP approach. Cross-Conformal Prediction, which was 
recently proposed in~\cite{vovk:ccp}, tries to overcome this problem by combining ICP with cross-validation. 
Specifically, CCP partitions the training set in $K$ subsets (folds) $S_1, \dots, S_K$ and calculates 
the nonconformity scores of the examples in each subset $S_k$ and of $(x_{l+1}, Y_j)$ for each possible 
classification $Y_j$ as
\begin{subequations}\label{eq:nmccpdef}
\begin{align}
\alpha_i &= A(\cup_{m \neq k} S_m, (x_i,y_i)), \mspace{15 mu} i \in S_k, \mspace{15 mu} m = 1, \dots, K, \\
\alpha^{Y_j,k}_{l+1} &= A(\cup_{m \neq k} S_m, (x_{l+1},Y_j)), \mspace{15 mu} m = 1, \dots, K,
\end{align}
\end{subequations}
where $A$ is the given nonconformity measure. Note that for $(x_{l+1},Y_j)$ $K$ nonconformity scores 
$\alpha^{Y_j,k}_{l+1}$, $k = 1, \dots, K$ are calculated, one with each of the $K$ folds.
Now the p-value for each possible classification $Y_j$ is computed as
\begin{equation}
\label{eq:pvalueccp}
  p(Y_j) = \frac{\sum^K_{k=1} |\{(x_i,y_i) \in S_k : \alpha_i \geq \alpha^{Y_j,k}_{l+1}\}| + 1}{l+1}.
\end{equation}

The CCP version of the framework was chosen to be followed here due to its big advantage in computational 
efficiency over CP, since it needs to train the underlying algorithm only $K$ times, and its advantage over 
ICP since it utilizes the whole training set for producing its p-values. 
As opposed to CP and ICP, the validity of which has been proven theoretically, at the moment there are no 
theoretical results about the validity of CCP. However in~\cite{vovk:ccp} its outputs have been shown to be 
empirically valid. The same is shown in the experimental results of this work, presented in Section~\ref{sec:res}.

\section{ML-RBF Cross-Conformal Predictor}\label{sec:nm}

This section describes the proposed approach, which in effect comes down to the definition of a suitable 
nonconformity measure for multi-label classification and its use with the CCP framework. In order to use the
CCP framework in the multi-label setting, the set of possible classifications $\{Y_1, \dots, Y_c\}$ is replaced 
by the powerset $\mathcal P\left({\{Y_1, \dots, Y_n\}}\right)$, where $n$ is the number of the original classes 
of the problem. In the 
experiments that follow RBF Neural Networks for multi-label learning (ML-RBF) \cite{zhang:mlrbf} is used as 
underlying algorithm as it seems to be one of the best performing algorithms designed specifically 
for multi-label problems. However the proposed approach is general and can be used with any other 
method which gives scores for each class.

After being trained on a given training set, for every test example ML-RBF produces a score for each 
possible class of the task at hand. It then outputs as its prediction the labelset containing all classes with 
score higher than zero; higher score indicates higher chance of the class to be in the labelset. 
In order to use the scores produced by ML-RBF for computing the nonconformity scores 
of the proposed Cross-Conformal Predictor, the former were transformed to the range $[0, 1]$ with the 
logistic sigmoid function
$$
f(x) = \frac{1}{1 + e^{-x}}.
$$
The nonconformity measure (\ref{eq:nmccpdef}) for the multi-label CCP can now be defined based on the transformed 
outputs of ML-RBF for $(x_i, \psi_i)$ after being trained on $\cup_{m \neq k} S_m$, $m = 1, \dots, K$ as
\begin{equation}
\label{eq:nm}
  \alpha_i = \sum^n_{j=1} |t^j_i - o^j_i|^d,
\end{equation}
where $n$ is the number of possible classes, $t^j_i$ is 1 if $Y_j \in \psi_i$ and 0 otherwise, 
and $o^j_i$ is the transformed output of the ML-RBF corresponding to class $Y_j$. Finally $d$ is a parameter of the 
algorithm which controls the sensitivity of the nonconformity measure to small differences in $o^j_i$ when 
$|t^j_i - o^j_i|$ is small as opposed to when it is large. 
This nonconformity measure takes into account the distance of all outputs from the true values. Note that in the 
case of test examples $t^j_i$ is the corresponding value for each assumed labelset.

The nonconformity measure (\ref{eq:nm}) takes into account only the outputs of the ML-RBF for each instance. 
However, it is very strange to have a pair of labels in a labelset that have never appeared together in the 
training set. So (\ref{eq:nm}) was extended to take into account the occurrence of each pair of labels 
in the training set $\cup_{m \neq k} S_m$, $m = 1, \dots, K$. This extended nonconformity measure is defined as:
\begin{equation}
\label{eq:nm2}
  \alpha_i = \sum^n_{j=1} |t^j_i - o^j_i|^d + \lambda\sum_{1\leq j<r \leq n} t^j_i t^r_i \mu_{j,r},
\end{equation} 
where $\mu_{j,r}$ is 0 if the labels $Y_j$ and $Y_r$ have been observed together in the labelset of at least one 
instance of the training set and 1 otherwise. In effect the additional part of this nonconformity measure 
adds $\lambda$ to the nonconformity score of an example for each pair of labels in the labelset of the example 
($t^j_i t^r_i = 1$) which have never been observed together in any instance of the training set ($\mu_{j,r} = 1$). 
The parameter $\lambda$ adjusts the sensitivity to this part of the measure. A high value of $\lambda$ 
makes this part of the measure dominate when a pair of labels that has never been observed before exists 
in the labelset of the example.

The complete proposed approach is derived by plugging in (\ref{eq:nm2}) as $A$ in (\ref{eq:nmccpdef}) and 
computing the p-value of each possible labelset with (\ref{eq:pvalueccp}). The resulting p-values can be used 
in either of the two ways described in Section \ref{sec:cp}.

\section{Experiments and Results}\label{sec:res}

\subsection{Data Sets}

To evaluate the performance of the proposed approach two data sets from different application domains were used, 
one from the semantic scene analysis domain and one from the bioinformatics domain. The first data set, 
\emph{scene} \cite{boutell:scene}, is concerned with the semantic classification of pictures into one or more 
of the classes: beach, sunset, foliage, field, mountain and urban. It consists of 1211 training and 1196 test examples, 
each described by 294 features. The second data set, 
\emph{yeast} \cite{Elisseeff:akernel}, is concerned with predicting the functional classes of genes in the 
Yeast Saccharomyces cerevisiae. 
Each gene is described by the concatenation of microarray expression data and a phylogenetic profile, and is associated 
with a set of 14 functional classes. The data set contains 1500 genes as training set and 917 genes as test set, 
each described by 103 features.
Both data sets were obtained from the website of the Mulan library \cite{Tsoumakas:miningmulti-label}.

\subsection{Single Prediction Evaluation}\label{sec:ressingle}

The first set of experiments evaluates the quality of the single predictions produced by the ML-RBF CCP and compares 
it to that of its underlying method and those of three other popular multi-label techniques, namely 
BP-MLL \cite{zhang:bpmll}, ML-kNN \cite{zhang:mlknn} and ML-NB \cite{zhang:mlnb}. Four evaluation measures for 
multi-label classification were used. The first is Hamming Loss (HL), which is the most popular measure for 
multi-label problems, defined as:
\begin{equation}
\label{eq:hloss}
  HL = \frac{1}{g} \sum^{l+g}_{i=l+1} \frac{|\psi_i \bigtriangleup \hat \psi_i|}{n},
\end{equation} 
where $\{(x_{l+1},\psi_{l+1}), \dots, (x_{l+g},\psi_{l+g})\}$ are the test examples, $\hat \psi_i$ is the predicted 
labelset for example $i$
and $\bigtriangleup$ is the symmetric difference between two sets. The second measure is Classification Accuracy (CA) 
defined as:
\begin{equation}
\label{eq:acc}
  CA = \frac{1}{g} \sum^{l+g}_{i=l+1} I(\psi_i = \hat \psi_i),
\end{equation}
where $I(true)=1$ and $I(false)=0$. This measure is rather strict as it requires the predicted and true labelsets to 
be identical. The third and fourth measures are the \emph{macro averaged} and \emph{micro averaged} 
F-measure, which is the harmonic mean of precision and recall. The F-measure for a single label is defined as:
\begin{equation}
\label{eq:fmeasure}
  F(tp,tn,fp,fn) = \frac{2tp}{2tp+fp+fn},
\end{equation}
where $tp$ is the number of true positives, $tn$ the number of true negatives, $fp$ the number of false positives and 
$fn$ the number of false negatives. In the multi-label case, if $tp_j$, $tn_j$, $fp_j$ and $fn_j$ are the same values 
for each label $Y_j$, then its macro averaged version is defined as:
\begin{equation}
\label{eq:macro}
  F_{macro} = \frac{1}{n} \sum^n_{j=1} F(tp_j,tn_j,fp_j,fn_j).
\end{equation}
The micro averaged version of the F-measure is defined as:
\begin{equation}
\label{eq:micro}
  F_{micro} = F \left(\sum^n_{j=1} tp_j, \sum^n_{j=1} tn_j, \sum^n_{j=1} fp_j, \sum^n_{j=1} fn_j \right).
\end{equation}

Tables~\ref{tab:pscene} and~\ref{tab:pyeast} present the performance of the proposed approach together with that of 
its underlying algorithm and that of BP-MLL, ML-kNN and ML-NB on the scene and yeast data sets
respectively. The best values for each measure are highlighted in bold. For ML-RBF the fraction parameter was set 
to 0.01 and the scaling factor to 1 as in \cite{zhang:mlrbf}. In the case of the CCP the number of folds for 
each data set was chosen 
so that each fold contained approximately 100 training instances, therefore 12 folds were used for the scene data set 
and 15 folds for the yeast dataset. The parameter $d$ of the nonconformity measure (\ref{eq:nm2}) was set 
to 4, which seems to be a good choice based 
on the performed experiments, while for $\lambda$ the two extreme values of 0 and 1 were used. Setting $\lambda$ to 0 
in effect corresponds to nonconformity measure (\ref{eq:nm}) and setting it to 1 makes the nonconformity score of 
any labelset containing a pair of classes that has never been observed in the training set always higher than others. 
For BP-MLL 
the number of hidden neurons was set to $20\%$ of the input dimensionality, the learning rate to 0.05 and the training 
epochs to 100 as in \cite{zhang:bpmll}. For ML-kNN the number of nearest neighbours was set to 10 and the smoothing 
parameter to 1 as in \cite{zhang:mlknn}. For ML-NB the percentage of remaining features after PCA was set to 0.3 as 
in \cite{zhang:mlnb}.

\begin{table}[t]
  \centering
  \caption{Performance of the proposed approach and comparison to that of other multi-label algorithms on the scene data set.}
  \label{tab:pscene}
  \begin{tabular}{@{\extracolsep{0.40cm}} l c c c c} \hline\noalign{\smallskip}
          & \multicolumn{4}{c}{Evaluation Measure} \\ \cline{2-5}\noalign{\smallskip}
Algorithm & HL & CA & $F_{macro}$ & $F_{micro}$ \\ \hline\noalign{\smallskip}
ML-RBF CCP with $\lambda = 0$ & 0.0928      & 0.6798      & \bf{0.7417} & \bf{0.7363} \\
ML-RBF CCP with $\lambda = 1$ & \bf{0.0927} & \bf{0.6831} & 0.7410      & 0.7358 \\
Orgiginal ML-RBF             & 0.0959      & 0.5468      & 0.6922      & 0.6890 \\
BP-MLL                       & 0.2903      & 0.1630      & 0.0509      & 0.1665 \\
ML-kNN                       & 0.0953      & 0.6012      & 0.7189      & 0.7183 \\
ML-NB                        & 0.1309      & 0.4105      & 0.6230      & 0.6221 \\ 
\hline
\end{tabular}
\end{table}

\begin{table}[t]
  \centering
  \caption{Performance of the proposed approach and comparison to that of other multi-label algorithms on the yeast data set.}
  \label{tab:pyeast}
  \begin{tabular}{@{\extracolsep{0.40cm}} l c c c c} \hline\noalign{\smallskip}
          & \multicolumn{4}{c}{Evaluation Measure} \\ \cline{2-5}\noalign{\smallskip}
Algorithm & HL & CA & $F_{macro}$ & $F_{micro}$ \\ \hline\noalign{\smallskip}
ML-RBF CCP with $\lambda = 0$ & \bf{0.1954} & 0.1821      & \bf{0.3896} & \bf{0.6432} \\
ML-RBF CCP with $\lambda = 1$ & \bf{0.1954} & 0.1821      & \bf{0.3896} & \bf{0.6432} \\
Orgiginal ML-RBF             & 0.1970      & \bf{0.1865} & 0.3891      & 0.6407 \\
BP-MLL                       & 0.2272      & 0.0960      & 0.3047      & 0.6212 \\
ML-kNN                       & 0.1980      & 0.1658      & 0.3567      & 0.6360 \\
ML-NB                        & 0.2115      & 0.1254      & 0.3428      & 0.6152 \\ 
\hline
\end{tabular}
\end{table}

The results in these tables show that not only the proposed approach provides important additional information 
about the likelihood of each of its predictions being correct, but it also outperforms its underlying algorithm 
and the three other popular multi-label techniques. The only exception is the classification accuracy of the ML-RBF 
in the case of the yeast data set, but even in this case the accuracy of the proposed approach with both values of 
$\lambda$ is very close. Comparing the performance of the ML-RBF CCP with the two different $\lambda$ values one can 
see that in the case of the scene data there is only a negligible difference while in the case of the yeast data the 
performance remains the same for all evaluation measures. Therefore the value of $\lambda$ does not have any important 
effect on the performance of the single predictions produced by the ML-RBF CCP. It does however affect the quality of 
the resulting p-values as will be shown in the next subsection.

\subsection{Prediction Region Evaluation}

The main advantage of the proposed approach over other multi-label techniques is the production of a p-value for 
each possible labelset of a new unseen instance, which can be translated either to confidence and credibility 
measures for its prediction or to prediction sets that are guaranteed to contain the true labelset at a required 
confidence level. This subsection examines the informativeness and reliability of the resulting prediction sets and 
consequently of the computed p-values and confidence measures. More specifically 
given a required level of confidence $1-\delta$, the ML-RBF CCP produces a set of labelsets that has at most 
$\delta$ chance of not containing the true labelset of the unseen instance. The informativeness of this set of 
labelsets can be assessed in terms of its size, while its reliability can be assessed by the percentage of cases 
for which it does not contain the true labelset, this percentage should be less than or very near $\delta$.

Tables \ref{tab:rscene} and \ref{tab:ryeast} present the results of the proposed approach in this setting for the scene 
and yeast data sets respectively with the nonconformity measure parameter $\lambda$ set to 0 and 1. The 
same parameters reported in Subsection~\ref{sec:ressingle} were used. The two tables report the sizes 
of the prediction sets produced for the 95\%, 90\% and 80\% confidence levels together with the observed error 
percentages, i.e. the percentages of prediction sets that did not contain the true labelset. 

\begin{table}[t]
  \centering
  \caption{Prediction set sizes and error rates at the 95\%, 90\% and 80\% confidence levels for the scene data set.}
  \label{tab:rscene}
  \def\arraystretch{1.1}
  \begin{tabular}{|l|ccc|ccc|} \hline
                      & \multicolumn{3}{c|}{With $\lambda = 0$} & \multicolumn{3}{c|}{With $\lambda = 1$} \\
  \,\,\,\# of\,\,       & \multicolumn{3}{c|}{Confidence Level}   & \multicolumn{3}{c|}{Confidence Level} \\
  \,\,\,labelsets\,\, & 95\% & 90\% & 80\% &  95\% & 90\% & 80\% \\ \hline
  \,\,\,1\,\,         & \,\,\,\,\,\phantom{0}0.00\%\,\,\,\,\, & \,\,\phantom{0}2.34\%\,\, & \,\,\,\,\,54.93\%\,\,\,\,\, & \,\,\,\,\,\phantom{0}6.77\%\,\,\,\,\, & \,\,18.23\%\,\, & \,\,\,\,\,62.63\%\,\,\,\,\, \\
  \,\,\,2\,\,               & \phantom{0}0.08\% & 25.50\% & 34.20\% & \phantom{0}9.62\% & 32.69\% & 29.43\% \\
  \,\,\,3 to $2^2$\,\,   & 25.50\% & 65.64\% & 10.87\% & 43.31\% & 45.07\% & \phantom{0}7.94\% \\
  \,\,\,$(2^2+1)$ to $2^3$\,\, & 72.74\% & \phantom{0}6.52\% & \phantom{0}0.00\% & 40.05\% & \phantom{0}4.01\% & \phantom{0}0.00\% \\
  \,\,\,$(2^3+1)$ to $2^4$\,\, & \phantom{0}1.67\% & \phantom{0}0.00\% & \phantom{0}0.00\% & \phantom{0}0.25\% & \phantom{0}0.00\% & \phantom{0}0.00\% \\ \hline
  \,\,\,Errors\,\,          & \phantom{0}3.76\% & \phantom{0}9.28\% & 21.07\% & \phantom{0}3.60\% & \phantom{0}9.28\% & 20.99\% \\ \hline
\end{tabular}
\end{table}

\begin{table}[t]
  \centering
  \caption{Prediction set sizes and error rates at the 95\%, 90\% and 80\% confidence levels for the yeast data set.}
  \label{tab:ryeast}
  \def\arraystretch{1.1}
  \begin{tabular}{|l|ccc|ccc|} \hline
                    & \multicolumn{3}{c|}{With $\lambda = 0$} & \multicolumn{3}{c|}{With $\lambda = 1$} \\
  \,\,\# of\,\,     & \multicolumn{3}{c|}{Confidence Level}   & \multicolumn{3}{c|}{Confidence Level} \\
  \,\,\,labelsets\,\, & 95\% & 90\% & 80\% &  95\% & 90\% & 80\% \\ \hline
  \,\,\,$(2^6+1)$ to $2^7$\,\,       & \,\,\,\,\,\phantom{0}0.00\%\,\,\,\,\, & \,\,\phantom{0}0.00\%\,\, & \,\,\,\,\,\phantom{0}1.42\%\,\,\,\,\, & \,\,\,\,\,\phantom{0}0.00\%\,\,\,\,\, & \,\,\phantom{0}0.00\%\,\, & \,\,\,\,\,\phantom{0}1.42\%\,\,\,\,\, \\
  \,\,\,$(2^7+1)$ to $2^8$\,\,         & \phantom{0}0.00\% & \phantom{0}0.11\% & \phantom{0}3.71\% & \phantom{0}0.00\% & \phantom{0}0.22\% & \phantom{0}4.58\% \\
  \,\,\,$(2^8+1)$ to $2^9$\,\,         & \phantom{0}0.55\% & \phantom{0}3.82\% & 14.07\% & \phantom{0}0.76\% & \phantom{0}4.36\% & 18.65\% \\
  \,\,\,$(2^9+1)$ to $2^{10}$\,\,      & \phantom{0}3.05\% & \phantom{0}7.09\% & 60.96\% & \phantom{0}3.93\% & 10.14\% & 60.32\% \\
  \,\,\,$(2^{10}+1)$ to $2^{11}$\,\,   & \phantom{0}8.94\% & 34.46\% & 19.85\% & 13.85\% & 45.58\% & 15.05\% \\
  \,\,\,$(2^{11}+1)$ to $2^{12}$\,\,   & 38.71\% & 52.89\% & \phantom{0}0.00\% & 57.58\% & 39.48\% & \phantom{0}0.00\% \\
  \,\,\,$(2^{12}+1)$ to $2^{13}$\,\,   & 48.64\% & \phantom{0}1.64\% & \phantom{0}0.00\% & 23.88\% & \phantom{0}0.22\% & \phantom{0}0.00\% \\
  \,\,\,$(2^{13}+1)$ to $2^{14}$\,\,\, & \phantom{0}0.11\% & \phantom{0}0.00\% & \phantom{0}0.00\% & \phantom{0}0.00\% & \phantom{0}0.00\% & \phantom{0}0.00\% \\ \hline
  \,\,\,Errors\,\,                  & \phantom{0}4.69\% & \phantom{0}9.38\% & 19.85\% & \phantom{0}4.80\% & \phantom{0}9.60\% & 19.96\% \\ \hline
\end{tabular}
\end{table}

Table \ref{tab:rscene} reports the results for the scene data set in terms of the percentage of prediction sets containing 
only 1, 2, 3 to 4, 5 to 8 and 9 to 16 labelsets for each confidence
level; there was no prediction set containing more than 16 labelsets out of the possible 63. 
The last row of the table reports the percentage of errors observed for each confidence level.
Comparing the results obtained with the 
nonconformity measure parameter $\lambda$ set to 0 with those 
obtained with $\lambda = 1$, one can see that the latter produces much more informative prediction sets. 
Taking into account the classification accuracy of this data set ($68.31\%$) 
and the large number of possible labelsets, the resulting prediction sets are quite tight. One can be 95\% confident 
in about 60\% of the test instances by 
considering less than 4 out of the possible 63 labelsets. By reducing the required confidence to 90\% one can be 
certain in a single labelset for about 18\% of the test instances and between one or two labelsets for about half the 
test instances. Finally with a confidence level of 80\% a single labelset is given for more than 60\% of the test 
instances. In terms of empirical reliability, only the percentages of errors for the 80\% confidence level are slightly 
higher than the required significance level, which can be attributed to statistical fluctuations.

Table \ref{tab:ryeast} presents the same results for the yeast data set. In this case the prediction sets contained 
a much higher number of labelsets so the table reports the percentage of prediction sets containing between $2^i+1$ 
and $2^{i+1}$ labelsets with $i = 6, \dots, 13$ for each confidence level; there were no prediction sets containing 
less than $2^6$ labelsets. The rather big size of the resulting prediction sets is not strange baring in mind the 
very low classification accuracy of this data set, which is only 18.65\%. Comparing the results obtained with the 
nonconformity measure parameter $\lambda$ set to 0 with those obtained with $\lambda = 1$, again shows the superiority 
of nonconformity measure (\ref{eq:nm2}), as it produces smaller prediction sets. Considering the high difficulty 
of the particular task one can say that the resulting prediction sets are quite informative. The number of labelsets 
needed to satisfy the 80\% confidence level is $1/16$th or less ($\leq 2^{10}$) of all the possible labelsets for 85\% of the 
test instances. Finally in terms of empirical reliability, the percentage of errors observed is in all cases below 
the required significance level.

\section{Conclusions}\label{sec:conc}

This work examined the application of the conformal prediction framework to the multi-label setting. Unlike the other 
techniques developed for multi-label problems, the proposed approach accompanies each of its predictions with reliable 
measures of confidence. Experimental results on two 
popular multi-label data sets have shown that not only the proposed approach provides important additional information 
for each prediction, but it also outperforms other popular multi-label techniques. Furthermore 
its confidence measures have been shown to be informative and reliable. The provision of confidence measures can 
be very helpful in practical applications, considering the high uncertainty that exists in this setting.

Future work includes the development of additional nonconformity measures and the experimentation with more 
multi-label data. In addition generating separate p-values for each class and combining them for obtaining the 
p-value of each labelset could also be examined as an alternative. Finally the possibility of generating a 
ranking of the possible classes for each instance would also be a good addition to multi-label CP.

\end{document}